\documentclass[runningheads]{llncs}

\usepackage{eccv}

\usepackage{eccvabbrv}

\usepackage{array}
\usepackage{adjustbox}
\usepackage{tabularray}
\usepackage{graphicx}
\usepackage{booktabs}

\usepackage[accsupp]{axessibility}  % Improves PDF readability for those with disabilities.

\usepackage[colorlinks]{hyperref}

\usepackage{orcidlink}

\usepackage{amsmath}
\begin{document}

% ---------------------------------------------------------------
% TODO REVIEW: Replace with your title
\title{Good Data Is All Imitation Learning Needs
% Fine-Tunning the Decision Boundary for DRL Agents
} 

% TODO REVIEW: If the paper title is too long for the running head, you can set
% an abbreviated paper title here. If not, comment out.
% \titlerunning{CF-IL}

% TODO FINAL: Replace with your author list. 
% Include the authors' OCRID for the camera-ready version, if at all possible.
\author{Amir Samadi\inst{1}\orcidlink{0009-0006-9399-7736} \and
Konstantinos Koufos\inst{1}\orcidlink{0000-0001-5653-2186} \and
Kurt Debattista\inst{1}\orcidlink{0000-0003-2982-5199} \and
\\Mehrdad Dianati\inst{1,2}\orcidlink{0000-0001-5119-4499}}

% TODO FINAL: Replace with an abbreviated list of authors.
\authorrunning{A.~Samadi et al.}
% First names are abbreviated in the running head.
% If there are more than two authors, 'et al.' is used.

% TODO FINAL: Replace with your institution list.
\institute{WMG, University of Warwick, UK \\
\email{amir.samadi, konstantinos.koufos, k.debattista, m.dianati@warwick.ac.uk} \and EEECS, Queen’s University Belfast, UK\\
\email{m.dianati@qub.ac.uk}
\\
\vspace{6pt}
Code: \href{https://github.com/Amir-Samadi/CF-IL}{\nolinkurl{github.com/Amir-Samadi/CF-IL}}}

% This research is sponsored by Centre for Doctoral Training to Advance the Deployment of Future Mobility Technologies (CDT FMT) at the University of Warwick.
% }

\maketitle

\begin{abstract}
In this paper, we address the limitations of traditional teacher-student models, imitation learning, and behaviour cloning in the context of Autonomous/Automated Driving Systems~(ADS), where these methods often struggle with incomplete coverage of real-world scenarios. To enhance the robustness of such models, we introduce the use of Counterfactual Explanations (CFEs) as a novel data augmentation technique for end-to-end ADS. CFEs, by generating training samples near decision boundaries through minimal input modifications, lead to a more comprehensive representation of expert driver strategies, particularly in safety-critical scenarios. 
This approach can therefore help improve the model's ability to handle rare and challenging driving events, such as anticipating darting out pedestrians, ultimately leading to safer and more trustworthy decision-making for ADS.
% To evaluate the efficacy of the proposed approach, CF-Driver, we fine-tuned a pre-trained model with the augmented dataset. 
Our experiments in the CARLA simulator demonstrate that CF-Driver outperforms the current state-of-the-art method, achieving a higher driving score and lower infraction rates. Specifically, CF-Driver attains a driving score of 84.2\%, surpassing the previous best model by 15.02 percentage points. These results highlight the effectiveness of incorporating CFEs in training end-to-end ADS. To foster further research, the CF-Driver code is made publicly available.

\end{abstract}

\section{Introduction}
\label{sec:intro}

\begin{figure}[!ht]
  \centering
  \begin{subfigure}{\linewidth}
    \centering
    \includegraphics[width=0.75\linewidth]{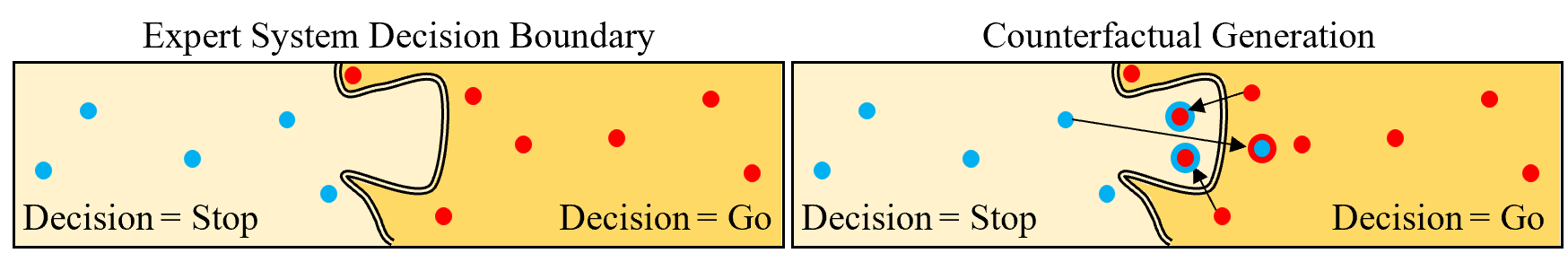}
  % \includegraphics[height=6.5cm]{img/1.jpg}
    % \fbox{\rule{0pt}{0.5in} \rule{.9\linewidth}{0pt}}
    \caption{Counterfactual sample generation involves minimally perturbing input data to cross the decision boundary, thereby changing the output.}
    \label{fig:teaser-a}
  \end{subfigure}
  % \hfill
  \\
  \begin{subfigure}{\linewidth}
    \centering
    \includegraphics[width=\linewidth]{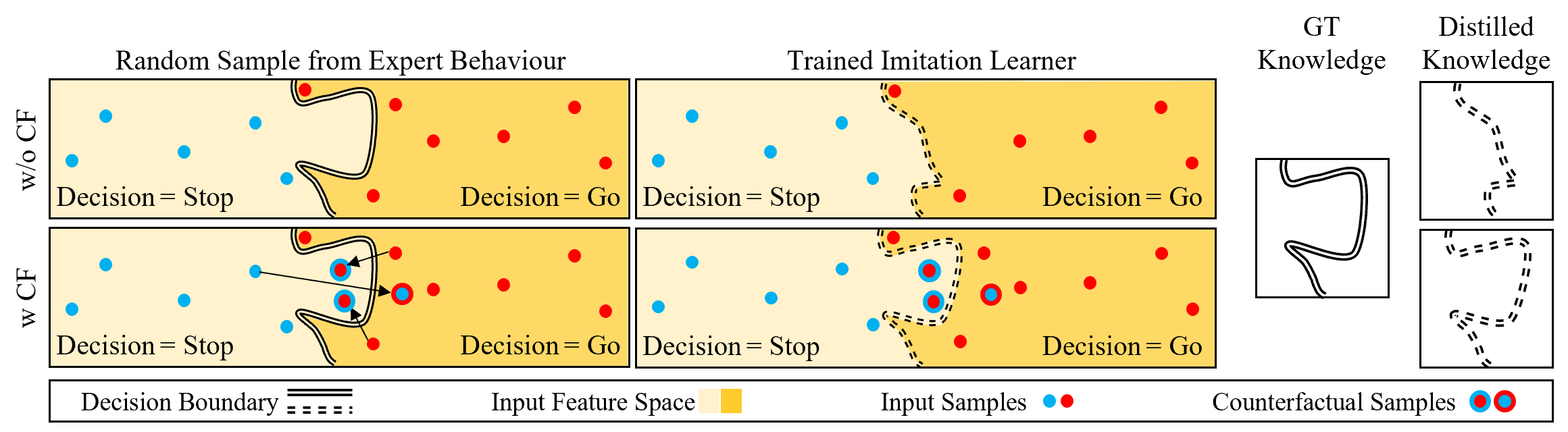}
    % \fbox{\rule{0pt}{0.5in} \rule{.9\linewidth}{0pt}}
    \caption{CF data is close to the decision boundary. Augmenting the sampled data with CFEs provides a more accurate representation of the decision boundary. This results in a closer similarity between the distilled decision boundary and the ground truth (GT) decision boundary, compared to distilled knowledge without CF samples (first row).}
    \label{fig:teaser-b}
  \end{subfigure}
  \\
  \begin{subfigure}{\linewidth}
    \centering
    \includegraphics[width=\linewidth]{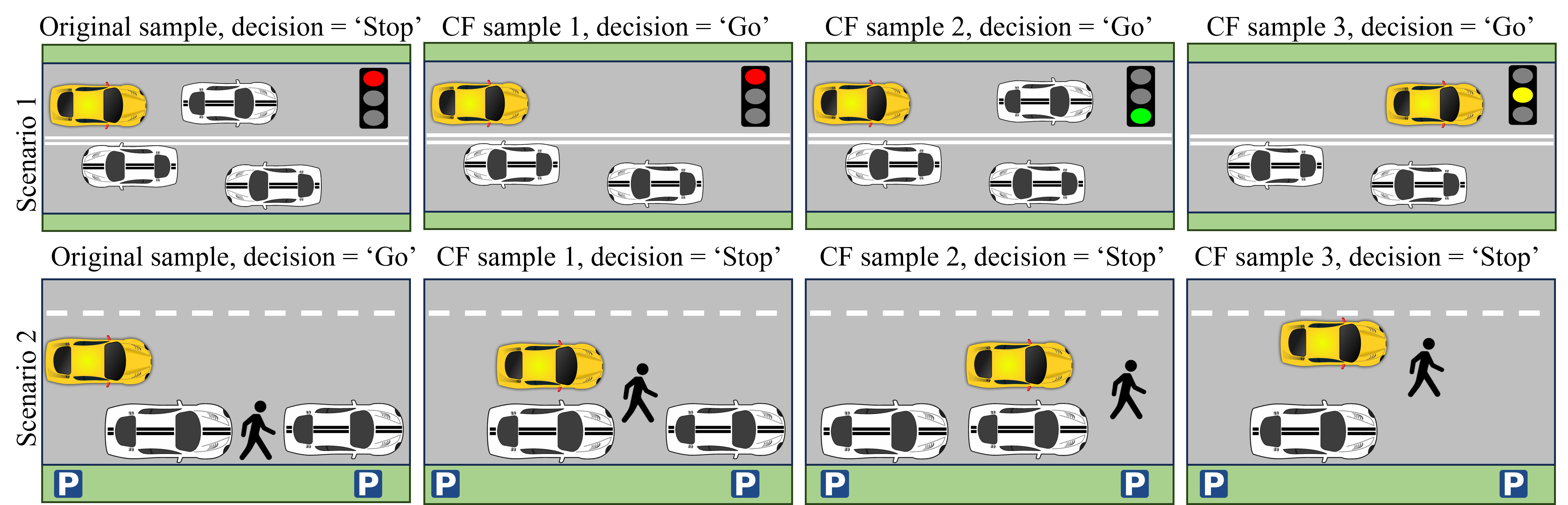}
    % \fbox{\rule{0pt}{0.5in} \rule{.9\linewidth}{0pt}}
    \caption{Two scenarios illustrate how CF samples enrich random sampling for a comprehensive understanding of an expert driver's strategy. Scenario 1: 
    % CF samples clarify the expert's stop decision for ego-vehicle (yellow car) due to a stopped vehicle ahead (white car).\\
    CF samples help the imitation learner understand that the expert driver's stop decision in the original sample stems from the stopped PV (white car) in front of the EV (yellow car).
    Scenario 2: CF samples highlight rare occluded pedestrian cases, leading to a more cautious driving policy near parked cars. CF samples enable the imitation learner to develop a robust understanding of the expert's decision-making process in various driving situations.
    }
    \label{fig:teaser-c}
  \end{subfigure}
  \\
  % \begin{subfigure}{0.68\linewidth}
  %   \includegraphics[height=2.5cm]{img/3.jpg}
  %   % \fbox{\rule{0pt}{0.5in} \rule{.9\linewidth}{0pt}}
  %   \caption{Another example of a subfigure}
  %   \label{fig:teaser-b}
  % \end{subfigure}
  \caption{Data Augmentation with CFEs for Improved Knowledge Distillation.}
  \label{fig:teaser}
\end{figure}

The rapid advancement of autonomous driving technologies has brought the challenges of dynamic scene understanding, situation awareness and decision-making to the forefront of research~\cite{giunchiglia2023road}. In the design of End-to-End (E2E) ADS (also referred to as the AV2.0 industry), researchers strive to automatically distill the intelligence of expert drivers—particularly their situational awareness and decision-making capabilities—into machine learning models. Techniques such as teacher-student models, imitation learning, and behaviour cloning have become popular for their ability to transfer knowledge from expert systems to student or clone models~\cite{jia2023driveadapter}. These methods typically involve training a model to replicate the decision-making process of an expert by learning from datasets of expert actions paired with sensory inputs. However, these approaches often struggle to capture the full spectrum of real-world scenarios, especially the rare but critical events that occur in the long tail of the distribution~\cite{corrado2023guided, ankile2024juicer}. This limitation, which is often referred to as uncurated data problem~\cite{belkhale2024data}, can lead to the transfer of sub-optimal behaviours, thereby limiting the model's performance in safety-critical applications such as those encountered in autonomous driving~\cite{shao2023reasonnet}.

To overcome such limitations, this paper proposes the use of Counterfactual Explanations~(CFEs)~\cite{wachter2017counterfactual} as a novel data augmentation technique within the context of E2E automated driving. CFE is an established method in Explainable AI~(XAI), designed to generate examples close to decision boundaries by making minimal changes to the input data that alter the model’s output (see Fig.~\ref{fig:teaser-a}). By incorporating CFEs into the training process, we aim to provide a more comprehensive representation of expert driver strategies, particularly in scenarios where traditional methods struggle.

The inclusion of CFEs enriches the training datasets with examples that better capture the nuances of expert decision-making, especially in challenging situations that are underrepresented in the original training data. Fig.~\ref{fig:teaser-b} demonstrates the enriching process, where incorporating CFEs helped the imitation learner model in inferring a decision boundary (distilled knowledge) similar to the ground truth (GT knowledge) expert boundary. For instance, in Fig.~\ref{fig:teaser-c} we illustrate two scenarios where the CF samples: (1) Help the imitation learner understand that the primary reason for the decision to stop is initiated by the stationary Participant Vehicle~(PV) in front. For moving PV or no PV in front of the Ego Vehicle~(EV), the latter can stop closer to the traffic light (when it is red), or keep driving when it is green or yellow. (2) Enrich the rare case of an occluded pedestrian by expanding the samples of pedestrians prohibiting the movement of the EV. The imitation learner model can link these images and interpret the possibility of an occluded object between parked cars, thus implementing a more cautious driving manoeuvre near parked vehicles. By incorporating CF samples, the imitation learner can develop a more robust and comprehensive understanding of the expert driver's decision-making process in various driving situations. This enhancement is crucial for improving the trustworthiness of E2E ADS. % ultimately contributing to safer and more reliable decision-making on the road.

In this paper, we propose a novel E2E CARLA driver, called CF-Driver, which creates rare and edge-case driving scenes through the generation of CF samples. That allows existing driver models to be fine-tuned to reach a driving strategy nearer to the expert driving policy. To evaluate the performance of CF-Driver, we have fine-tuned the second-best model on the CARLA leaderboard~\cite{carla-leaderboard}, named the Interfuser model~\cite{shao2023safety}, within our proposed framework, and achieved superior results over the top-ranking model in the CARLA leaderboard. To foster further research studies, we made the generated datasets publicly available, encompassing multi-modal sensor inputs (RGB Camera, Lidar, GNSS, IMU and Speedometer) labelled with an expert driver and enriched with CF samples. In summary, the key contributions of this paper are:    
\begin{enumerate}
    \item 
    The proposition of CF-Driver, which leverages counterfactual examples to capture a comprehensive expert driving policy, resulting in State-of-the-Art (SOTA) performance in the CARLA simulator.
    \item A comprehensive performance evaluation analysis of CF-Driver with eight SOTA CARLA driver models and a demonstration of superior performance.  
    \item The generation of new enriched datasets for multiple city environments in CARLA labeled by an expert driver.
\end{enumerate}

 \section{Background and Related work}
 \label{section:related}
 This section provides an overview of the existing literature in the realm of interpretability, focusing on the use of counterfactual explanations~(CFEs) as a promising approach to augment datasets. We begin by briefly reviewing CFE generating approaches and the studies that have been conducted to evaluate CFE capability in enriching data. Then, we review the state-of-the-art CARLA drivers, especially those that have used Imitation Learning~(IL) methods.

\subsection{Counterfactual Explanations}
\label{sec:RW-CF}
CFEs have gained significant attention in the field of explainable AI (XAI) due to their ability to provide insight into the decision-making process of machine learning models~\cite{jacob2022steex, samadi20203SAFE}. A CFE describes a minimal change in the input features that would alter the model's output~\cite{wachter2017counterfactual}. By generating such examples, researchers aim to identify the critical factors influencing the model's decisions and provide a more comprehensive understanding of its behaviour~\cite{samadi2023counterfactual}.

Several approaches have been proposed to generate CFEs. Wachter~\emph{et al.}\cite{wachter2017counterfactual} introduced the concept of CFEs and proposed a gradient-based method to generate them. Following that, other optimisation methods such as genetic algorithms~\cite{dandl2020multi}, game theory~\cite{ramon2020comparison}, and Monte Carlo methods~\cite{lucic2020does} have been proposed. Mothilal~\emph{et al.}~\cite{mothilal2020dice} extended this work by using a diverse CF framework that generates multiple CFEs for a given input, which we have used in this study. While the above-mentioned methods can only generate CFEs for low-dimensional inputs such as tabular data, Mahajan~\emph{et al.}~\cite{mahajan2019preserving} introduced a generative modeling approach to produce CFEs to provide CFEs for high-dimensional inputs such as images leveraging the power of deep learning. However, they can introduce artefacts along the generated CFEs, which has been discussed in~\cite{samadi2024safe}. 

CFEs are not only used to provide insight into black-box deep neural networks~(DNNs) but also several studies have shown their effectiveness in enriching datasets and improving model performance~\cite{joshi2021investigation}. Teney~\emph{et al.}~\cite{teney2020learning} demonstrated that incorporating CFEs into the training process of visual question answering models led to improved accuracy and robustness. Goyal~\emph{et al.}~\cite{goyal2019counterfactual} used CFEs to augment datasets for facial expression recognition, resulting in better generalisation and reduced bias. Recent studies have shown the effectiveness of incorporating human guidance and counterfactual data augmentation in IL. Corrado~\emph{et al.}~\cite{corrado2023guided} proposed the Guided Data Augmentation (GuDA) framework, which provides expert-quality data by focusing on task-advancing augmentations. Ankile~\emph{et al.}\cite{ankile2024juicer} introduced a pipeline that improves IL performance in complex assembly tasks by integrating advanced policy architectures with dataset expansion and simulation-based data augmentation techniques. Sun~\emph{et al.}~\cite{sun2023offline} proposed the Offline IL with Counterfactual Data Augmentation (OILCA) framework, which enhances agent performance and generalisation by generating high-quality expert data using counterfactual inference.

\subsection{CARLA Drivers and Imitation Learning}
\label{sec:RW-IL}
CARLA~\cite{dosovitskiy2017carla} is a popular open-source simulator for Automated Driving Systems~(ADS) research, providing a realistic environment for training and evaluating self-driving models. Several SOTA CARLA drivers have been developed using IL techniques~\cite{chen2020learning} aiming to learn a driving policy by mimicking expert demonstrations. IL is a powerful approach for teaching autonomous agents to perform tasks by learning from expert domain knowledge~\cite{hussein2017imitation, jaeger2021expert}. In the context of ADS, IL involves training a model to replicate the behaviour of an expert driver using a dataset of expert actions paired with corresponding sensory inputs~\cite{chen2021learning}. Teacher-student models and behavioural cloning are two common IL techniques used in developing ADS.

Teacher-student models, also known as knowledge distillation, involve training a student model to mimic the behaviour of a pre-trained teacher model~\cite{hinton2015distilling}. The teacher model, which is typically a large and complex network, is first trained on a large dataset of expert demonstrations. Then, the student model, which is usually a smaller and more efficient network, learns to reproduce the teacher's outputs using a subset of the training data. This approach allows for the transfer of knowledge from the teacher to the student, resulting in a more compact and efficient model suitable for real-time inference in autonomous vehicles~\cite{samadi2023counterfactual}.

Behavioural cloning directly learns a mapping from sensory inputs to control commands by training a model to minimise the difference between its predicted actions and the expert's actions~\cite{codevilla2018end}. The model is typically a DNN that takes in sensory data such as camera images, LiDAR point clouds \cite{haghighi2024taming} and GPS coordinates, and outputs control commands such as steering angle, throttle, and brake. Behavioural cloning has been successfully applied to learn E2E driving policies in CARLA~\cite{codevilla2018end, liang2018cirl}.

Conditional IL (CIL)~\cite{codevilla2018end} is a widely adopted approach in CARLA, where a DNN is trained to map sensory inputs to control commands, conditioned on high-level navigation commands. CIL has been extended and improved in various ways, such as incorporating attention mechanisms~\cite{zhang2019learning} and using adversarial training~\cite{kuefler2017burn} to enhance the robustness of the learned policy. Other notable CARLA drivers include the Reinforcement Learning~(RL)-based approach by Liang~\emph{et al.}~\cite{liang2018cirl}, which combines IL with RL to improve the driving performance, and the Vision Transformer-based model by Prakash~\emph{et al.}~\cite{prakash2021multi}, which leverages the power of transformers to learn a more effective representation of the driving scene.

Despite the success of IL in developing ADS, these approaches often struggle with distributional shift~\cite{ross2011reduction}. Distributional shift occurs when the distribution of scenarios encountered during testing differs from the training data distribution~\cite{sagawa2019distributionally}. This can happen due to the limited coverage of expert demonstrations, which may not capture all possible driving situations. As a result, the learned driving policy may not generalize well to novel scenarios, leading to suboptimal or unsafe behaviour. To address the limitations of IL, researchers have proposed various techniques to improve the robustness and generalisation of learned driving policies~\cite{sagawa2019distributionally}. One approach is to combine IL with RL, allowing the agent to learn from experience and adapt to new situations~\cite{wang2016dueling, liang2018cirl}. Another approach is to incorporate uncertainty estimation and risk-aware planning into the decision-making process, enabling the agent to handle uncertain and ambiguous situations more effectively~\cite{michelmore2020uncertainty, filos2020can}.

In this study, we propose a novel data augmentation framework using CFEs to enrich the training data for IL-based CARLA drivers. By generating counterfactual examples that capture rare and critical events, we aim to improve the coverage of the training data and enhance the robustness of the learned driving policy. Our approach complements existing techniques and contributes to the development of safer and more reliable ADS.

 \section{Methodology}
 \label{section:method}
 
\begin{figure}[tb]
  \centering
  \includegraphics[width=\linewidth]{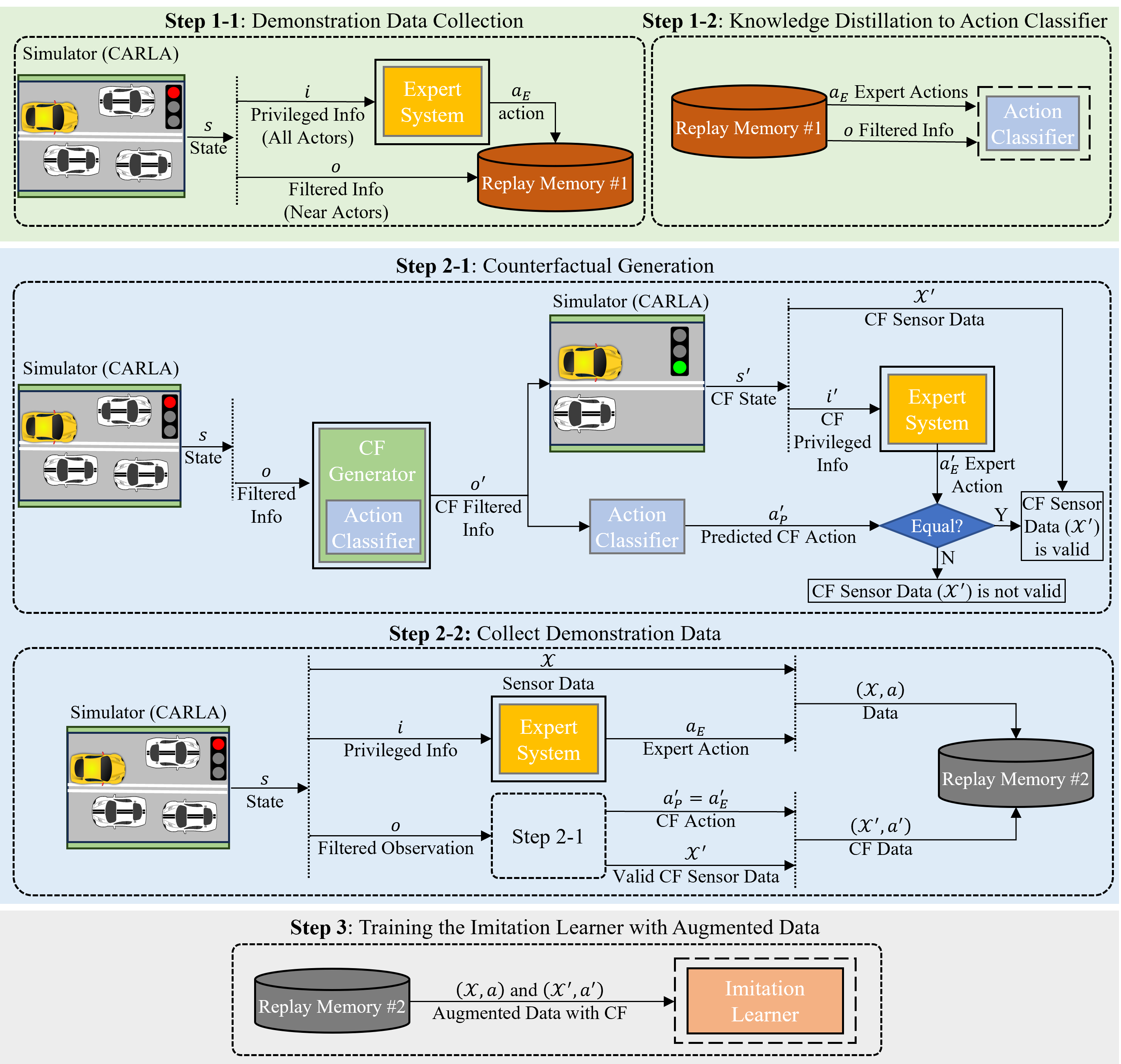}
  \caption{Overview of the three-step process used to train the proposed CF-Driver. So
  lid lines surrounding the model boxes indicate fixed models, while dashed lines represent models undergoing training.}
  \label{fig:method}
\end{figure}

This section describes the proposed CF-Driver framework, which enhances Imitation Learning (IL) efficiency by enriching the demonstration dataset. We begin by defining the problem and then proceed with the details of the CF-Driver framework.

\subsection{Problem Definition}

The CF-Driver is based on behaviour cloning, where an agent learns to imitate an expert policy $\pi^{*}$. Following the approach described in~\cite{prakash2021multi, chitta2021neat}, the expert driver at each time step $t$ uses the privileged simulator information $i$ (for clarity, the subscript $t$ is omitted hereafter), including all the actors' states to generate the imminent future trajectory $a_E$ for the Ego Vehicle~(EV). The road actor states encompass location, rotation and velocity for dynamic actors, and traffic-light status. Given a known EV model, the imminent future trajectory denoted by $a$, includes the waypoints, brake, and acceleration signals for the next $10$ time steps. A waypoint is defined on the Bird's Eye View~(BEV) plane as a set of 2D coordinates $w = (x, y)$. The low-level lateral and longitudinal controllers in CARLA use PID with the lateral controller minimising the EV angle to the next waypoints, and the longitudinal controller maintaining a fixed target speed or stopping. 
% The vehicle stops to obey traffic rules or avoid collisions, with risks detected by identifying obstacles in a designated area ahead.

The expert driver processes all actor information $i$ and provides actions, including waypoints, acceleration and brake commands, for the next $10$ time steps $a_E = (w_E, v_E, b_E)$. Thus, the Expert can be defined as $\pi^{*}(i) = a_E$. Our objective is to train an Imitation Learner $\pi^{L}$ that receives multimodal sensor input data $\mathcal{X}$ (instead of the privileged info $i$) and effectively mimics the Expert driving policy, i.e., $\pi^{L}\approx\pi^{*}$.

\subsection{Counterfactual Driver}
CF-Driver navigates frame-by-frame in the Carla urban map, completing given routes while avoiding collisions and obeying traffic rules. The goal is to train an agent that processes multi-modal sensor inputs (RGB Camera, Lidar, GNSS, Radar, IMU and Speedometer) to develop a driving policy $\pi$ that follows predefined routes. Fig.~\ref{fig:method} illustrates the proposed framework that comprises three main steps: (1) Data collection from an expert driving heuristic algorithm and training a tree-based student model, (2) collection of expert and Counterfactual (CF) sample data, and (3) training an Imitation Learner using both CF and regular sample data. The following sections describe each proposed step in detail.
\subsubsection{Data Collection and Tree-based Classifier:}
In the first step of the proposed CF-driver (refer to Step~1-1 and Step~1-2 in Fig.~\ref{fig:method}), we train the student model to act similarly to the Expert System. The reason for training a student model lies in the fact that the expert driver is a rule-based model that is not differentiable, while to generate the CF dataset, we need a differentiable model to minimise the CF loss function defined in Eq.~\eqref{eq:eq1}. The clone student model will then be used by a CF generator model to provide CF data. To reduce the computational complexity of the student model, we first filter the extensive information $i$ retaining only the state of the nearest four actors to the EV (could be pedestrians, cyclists, vehicles or traffic lights). This includes their relative location, speed, and rotation relative to the EV, denoted by $o$, and associated with the expert action. This data is used to train a Tree-based action classifier~$\pi^T(o)=a_P$, which learns to mimic the Expert bahaviour~$\pi^T\approx\pi^*$.

\subsubsection{CF Example Generation:}
In Step~2-1 of the proposed framework, the trained action classifier (obtained by Step~1) is utilised within the CF generator module to provide CFEs $o^{\prime}$. 
%Given an original input and output, a CF example is the minimally perturbed input that changes the original output. 
According to the definition of CFEs, we apply minimal alterations to the filtered observation $o$ to generate a CF observation $o^{\prime}$ that changes the output of the Expert and cloned tree-based action classifier to $a^{\prime}$ such that $ a^{\prime} = \pi^*(i^{\prime}) = \pi^T(o^{\prime})\neq a $. To obtain such CF examples, we employ the following loss function:
\begin{equation}
\label{eq:eq1}
\mathcal L_{CF}(o^\prime,\lambda;o,a,\pi^T)=\max_\lambda \lambda(\pi^T(o^\prime)-{a^\prime})^2 + \textnormal{dist}(o,o^\prime),
\end{equation}
where the coefficient term $\lambda>0$ balances the two terms in the loss function,  $\textnormal{dist}(\cdot,\cdot)$ denotes the Euclidean distance minimising the changes between the CF filtered state $o^\prime$ and the original filtered state $o$, and the subscripts $t$ from Eq.~\eqref{eq:eq1} have been omitted for presentation clarity. 

The loss function $\mathcal L_{CF}$ would be minimised when its first term $(\pi^T(o^\prime)-{a^\prime})^2$ becomes zero to alter the action from the original action $\pi^T(o)=a$, to the counter action $a^\prime\neq a$, and at the same time, the counter state $o^{\prime}$ remains near the decision boundary to minimise the distance $\textnormal{dist}(o,o^\prime)$. Intuitively, such sampled inputs can better represent the expert decision boundary for the Imitation Learner to understand and learn from. It is worth noting that we employed the DICE model~\cite{mothilal2020dice}, which generates multiple CFEs for a given input, further enriching the generated dataset (see Section~\ref{sec:RW-CF} for more details).

\subsubsection{Collect demonstration data:}
In Step 2-2, we deploy the Expert driver in the Carla Simulator environment for data augmentation. Within this environment and for each time step, we have access to multi-modal sensor data $\mathcal{X}$ used by the Imitation Learner model, privileged information $i$ used by the Expert, and filtered observations of the nearest actors' states $o$. At each time step, we store the sensor data $\mathcal{X}$ associated with expert trajectories $a_E$ as labels. Based on the CF generation method described under Step~2-1, we also generate CF states $\mathcal{X}^\prime$ that alter the trajectories outputted by the action classifier and expert system to $a^\prime$. In this way, the collected dataset $(\mathcal{X},a)$ generated by the Expert System within CARLA is enriched with the CF sensor data and their corresponding expert trajectories as labels $(\mathcal{X}^\prime,a^\prime)$.

\subsubsection{Training the Immitation Learner with Augmented Data:} Finally, in Step~3, we train the Imitation Learner model $\pi^L$ to mimic the expert driver. The enriched data set with the CF samples $(\mathcal{X}^{\prime},a^\prime)$ enables the Imitation Learner to better approximate the expert's decision boundary, resulting in improved adherence to traffic rules and safe driving, despite relying solely on sensor data. We base the Imitation Learner model on the InterFuser model proposed by Shao \etal~\cite{shao2023safety}. This model provides 10-point waypoints for steering and a prediction map for adherence to traffic rules. The prediction map includes 7 channels, providing information on potential grid occupation, traffic light status prediction, stop sign presence prediction, and prediction of whether the EV is located at a junction. The training objective is defined as:
\begin{equation}
\label{eq:eq2}
\mathcal{L} =\lambda_{pt} \mathcal{L}_{\text {pt}} + \lambda_{map} \mathcal{L}_{map} + \lambda_{tf}\mathcal{L}_{tf}, 
\end{equation}
where $\lambda_{pt}, \lambda_{map}$ and $\lambda_{tf}$ are coefficients balancing the loss components, $\mathcal{L}_{\text {pt}}$ is the waypoint prediction loss, $\mathcal{L}_{\text {map}}$ is the map prediction loss, and $\mathcal{L}_{tf}$ is the traffic rules and information loss. For the detailed descriptions of these loss functions, the reader may refer to~\cite{shao2023safety}.

 \section{Evaluations}
 \label{section:results}
 To evaluate the effectiveness of the CF-Driver and the impact of counterfactual explanations on imitation learning-based autonomous driving, we conducted a series of experiments using the CARLA simulator~\cite{dosovitskiy2017carla}. We compared the performance of CF-Driver with State-of-the-Art~(SOTA) CARLA drivers and analysed the benefits of incorporating counterfactual examples into the training data.

\subsection{Implementation Details}
\label{sec:ID}
We used CARLA version 0.9.10 for our experiments~\cite{dosovitskiy2017carla}, which provides a realistic urban driving environment with various weather conditions, traffic scenarios, and road layouts. We trained the models on the provided Town01-04 maps and tested them on the Town05, over long, short, and tiny scenario routes, which represent diverse driving scenarios. The expert demonstrations were sourced from the CARLA Expert driver~\cite{chitta2021neat}, which consists of high-quality driving data collected from a rule-based expert driver. We then trained an XGBoost action classifier to imitate the Expert Driver's behaviour. This action classifier was employed by the DICE~\cite{mothilal2020dice} counterfactual example generator model to augment a new CF-Expert dataset.
% For the expert demonstrations, we used the CARLA Expert driver~\cite{chitta2021neat}, which contains high-quality driving data collected from a rule-based expert driver. We augmented this dataset with counterfactual examples generated using the DICE framework~\cite{mothilal2020dice} to create the CF-Expert dataset.
For the Imitation Learner, we used the pre-trained Transformer Encoder-decoder model proposed in~\cite{shao2023safety}, named the Interfuser model. The hyper-parametrs in Eq.~\ref{eq:eq2} are set to: $\lambda_{pt}=0.4$, $\lambda_{pt}=0.4$ and $\lambda_{tf}=1$.

\subsection{Performance Metrics}
\label{sec: metric}
To evaluate the quality of the generated CF examples, we employed the following well-established CARLA Leaderboard~\cite{carla-leaderboard} metrics:
\begin{itemize}

\item Driving score: The primary metric used in the leaderboard, calculated as the average product of route completion percentage and infraction penalty across all routes. This metric provides a comprehensive evaluation of the agent's performance, considering both its ability to complete the assigned tasks and its adherence to traffic rules and safe driving practices.

\item Route completion: The percentage of successful episodes where the agent reaches the destination within the time limit (allocated by CARLA Leaderboard). This metric assesses the agent's ability to navigate and complete the assigned tasks effectively.

\item Infraction score: A measure of the agent's adherence to traffic rules and safe driving practices. It is calculated based on the number and severity of infractions committed during an episode, such as collisions, red light violations, and lane infractions. It functions as a performance deduction value. When the ego-vehicle violates a traffic rule or commits an infraction, the score decreases by a corresponding percentage. Higher infraction scores indicate safer and more rule-abiding driving behaviour.

% \item Vehicle, pedestrian, and layout collisions: The number of collisions between the agent's vehicle and other vehicles, pedestrians, or static objects in the environment. These metrics highlight the agent's ability to perceive and avoid obstacles, ensuring the safety of the agent and other road users.

% \item Red light violations: The number of times the agent fails to stop at red traffic lights. This metric evaluates the agent's compliance with traffic signals and its understanding of road rules.

% \item Offroad and blocked infractions: The number of times the agent drives off the designated road or becomes blocked by static objects. These metrics assess the agent's ability to stay within the drivable area and navigate around obstacles effectively.
 
\end{itemize}
These metrics provide a comprehensive assessment of the agent's driving performance, considering both task completion and safety aspects.

\subsection{Performance Evaluation}
\label{sec:PE}
To evaluate the effectiveness of our proposed CF-Driver, we compared its performance with eight SOTA CARLA drivers on the challenging Town05 benchmark, which features the longest and most complex routes. The results of this comparison are presented in Table~\ref{table:per}.
The CF-Driver achieves the highest driving score, which combines both route completion and infraction penalties,  showcasing its superior overall performance. That can be attributed to the incorporation of counterfactual explanation~(CFE) data during the fine-tuning process, which enables the CF-Driver to develop a more robust and safe driving policy.
In terms of route completion, CF-Driver outperforms all the SOTA models, successfully navigating a higher percentage of the challenging routes in Town05. This indicates that the inclusion of CFE data helps the model to better handle diverse and complex driving scenarios, resulting in improved navigation capabilities.
Moreover, CF-Driver exhibits lower infraction scores compared to the other methods, showcasing its ability to drive more safely and adhere to traffic rules. 
%The infraction score, which ranges from 0 to 1, takes into account various factors such as collisions, red light violations, and lane infractions. The higher infraction scores achieved by CF-Driver 
This demonstrates that the model has learned to make better decisions in critical situations, minimising risky behaviours and prioritising safety.
It is worth noting that the performance gains achieved by CF-Driver are consistent across all the individual metrics, as evidenced by the higher driving score. Therefore, the incorporation of CFE data not only improves the model's ability to complete routes but also enhances its understanding of safe driving practices.
% The results presented in Table~\ref{table:per} highlight the effectiveness of our proposed CF-Driver framework in learning a robust and safe driving policy. By leveraging CFE data to capture the nuances of expert behaviour in critical and rare scenarios, CF-Driver outperforms the SOTA models in terms of both route completion and infraction scores. This demonstrates the potential of counterfactual explanations in improving the performance and safety of autonomous driving systems.

\begin{table}[h]
\centering
\caption{ Comparison of the proposed CF-Driver with eight SOTA CARLA drivers on the longest routes in the Town05 benchmark. The driving score and road completion metrics are presented as percentages, while the infraction score ranges from 0 to 1, with higher values indicating better performance for all metrics. Our method outperforms the SOTA models not only in road completion but also in infraction scores, showcasing a safer driving strategy. The driving score metric summarises all the individual metrics, demonstrating the overall superiority of CF-Driver. }
\label{table:per}
\begin{tabular}{
>{\hspace{0pt}}m{0.25\linewidth}
>{\centering\hspace{0pt}}m{0.15\linewidth}
>{\centering\hspace{0pt}}m{0.15\linewidth}
>{\centering\arraybackslash\hspace{0pt}}m{0.15\linewidth}
} 
\toprule
Method & \begin{tabular}[c]{@{}c@{}}Driving \\ Score \end{tabular} & \begin{tabular}[c]{@{}c@{}}Road \\ Completion \end{tabular} & \begin{tabular}[c]{@{}c@{}}Infraction\\Score\end{tabular}  \\ 
\hline
CF-Driver~(Ours)                    & \textbf{84.23}&\textbf{96.23} & \textbf{0.88}\\
\hline
ReasonNet~\cite{shao2023reasonnet}  &73.22          & 95.88         & 0.76  \\
InterFuser~\cite{shao2023safety}    & 68.31          & 94.97         & 0.73\\
WOR~\cite{chen2021learning}         & 44.80          & 82.41         & 0.54\\
Roach~\cite{zhang2021end}           & 43.64          & 80.37         & 0.54\\
NEAT~\cite{chitta2021neat}          & 37.72          & 62.13         & 0.61\\
TransFuser~\cite{prakash2021multi}  & 33.15          & 56.36         & 0.59\\
LBC~\cite{chen2020learning}         & 7.05           & 32.09         & 0.22\\
CILRS~\cite{codevilla2018end}       & 3.68           & 7.19          & 0.51\\
\bottomrule
\end{tabular}
\end{table}

\begin{table}
\centering
\caption{An ablation study demonstrating the effectiveness of CFEs. The original pre-trained Interfuser model\cite{shao2023safety} and the fine-tuned Interfuser model using newly generated data without CFEs are compared against CF-Driver, which is the fine-tuned Interfuser model using newly generated data with CFEs. The performance of the expert driver is also included for reference. ``Compl.'', ``Infra.'', and ``Coll.'' represent completion, infraction and collision metrics, respectively. For the Driving Score, Route Completion, and Infraction Score metrics, higher values indicate better performance, while for the remaining metrics, lower values are preferred.}
\label{table:sota_detailed}
\adjustbox{width=\textwidth}{
\begin{tblr}{
  row{2} = {c},
  cell{1}{1} = {r=2}{},
  cell{1}{2-10} = {c},
  cell{3-6}{2-10} = {c},
  hline{1,7} = {-}{0.1em},
  hline{3} = {1}{0.03em},
  hline{3} = {2-10}{},
  column{1} = {l}{0.27\linewidth},
  column{2} = {c}{0.09\linewidth},
  column{3} = {c}{0.09\linewidth},
  column{4} = {c}{0.07\linewidth},
  column{5} = {c}{0.13\linewidth},
  column{6} = {c}{0.09\linewidth},
  column{7} = {c}{0.10\linewidth},
  column{8} = {c}{0.12\linewidth},
  column{9} = {c}{0.11\linewidth},
  column{10} = {c}{0.10\linewidth},
}
Method                                            & Driving & Route  & Infrac. & Pedestrian & Vehicle & Layout & Red light  & Offroad & Route   \\
                                                  & Score   & Compl. & Score   & Coll.      & Coll.   & Coll.  & Violations & Infrac. & Timeout \\
InterFuser~\cite{shao2023safety} & 72.31   & 93.61  & 0.76    & 0.00       & 0.05    & 0.00   & 0.08       & 0.01    & 0.02    \\
Fine-tunned InterFuser                            & 72.88   & 93.78  & 0.76    & 0.00       & 0.07    & 0.00   & 0.02       & 0.00    & 0.03    \\
CF-Driver~(Ours)                                  & 84.23   & 96.23  & 0.88    & 0.00       & 0.03    & 0.00   & 0.00       & 0.00    & 0.01    \\
Expert-Driver                                     & 86.92   & 99.14  & 0.87    & 0.01       & 0.03    & 0.00   & 0.00       & 0.00    & 0.02
\end{tblr}}
\end{table}
\subsection{Ablation study}
To better understand the impact of using CFE data in the CF-Driver framework, we conducted an ablation study by removing the CFE data and fine-tuning the Interfuser model with newly generated data that didn't include the CF scenarios (264246 original data points). As expected, the performance of the re-tuned model without CFE data is similar to that of the pre-trained Interfuser model (refer to Table~\ref{table:sota_detailed}). On the contrary, fine-tuning the model with the newly generated data together with the CF augmented data (234736 original and 32711 CF data points) significantly improved the driving score. This demonstrates that the CF-Driver has gained a better understanding of the expert driving decision boundaries, resulting in a safer driving policy. This is also evidenced by the similarity of the scores between the CF driver and the expert driver in Table~\ref{table:sota_detailed}.

Furthermore, the inclusion of CF data in the fine-tuning process also decreases the infraction penalty, indicating that the CF-Driver has learned to navigate the environment while committing fewer violations and adhering more closely to traffic rules. This reduction in infractions can be attributed to the CFE data, which provides the model with a more comprehensive understanding of the expert driver's decision-making process in critical and rare scenarios.

The ablation study highlights the importance of incorporating CFE data in the training process, as it enables the model to capture the nuances of expert behaviour and generalise better to unseen situations. By learning from a diverse set of scenarios, including those generated by counterfactual explanations, the CF-Driver develops a more robust and safe driving policy compared to models trained solely on the original data.

\subsection{Discussion}
\begin{figure}[h]
  \centering
  \includegraphics[width=\linewidth]{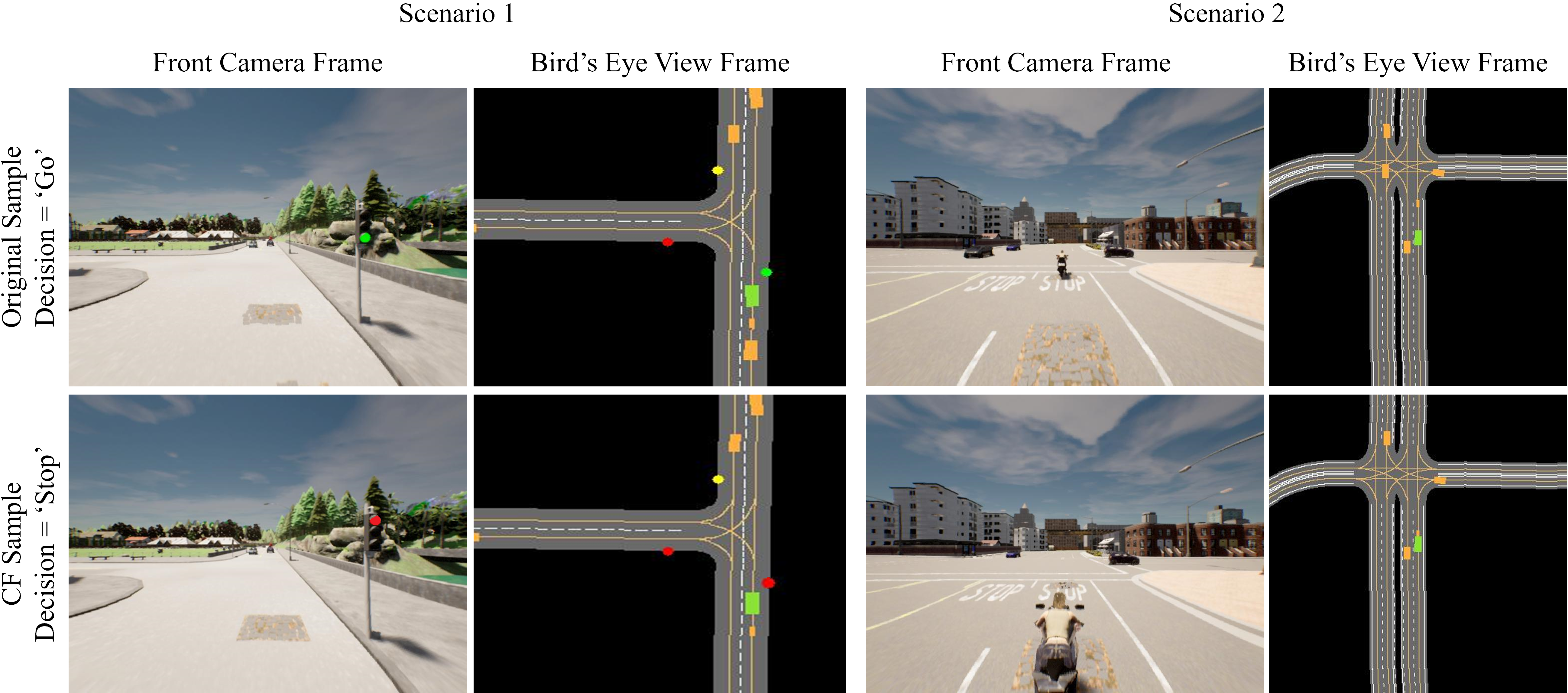}
  \caption{Illustration of the dataset consisting of front camera and Bird's Eye View frames, with original and counterfactual examples. In Scenario 1, the top row shows an original frame from Town01, where the vehicle is approaching a green traffic light. The bottom row demonstrates a counterfactual example generated by the CF generator, where the traffic light has been changed to red. The expert driver labels this counterfactual scenario as ``stop''. In Scenario 2, the CF generator modified the motorcycle's location, bringing it closer to the ego-vehicle. This change shifts the expert's decision from ``go'' to ``stop'', demonstrating how counterfactual examples can refine the imitation learner's decision boundary.}
  \label{fig:frames}
\end{figure}

Fig.~\ref{fig:frames} illustrates two scenarios from Town01 and Town03, demonstrating the effectiveness of our CFE generation approach. In the first scenario, while the original data shows the vehicle approaching a green traffic light, the CF generator provides an instance where the traffic light has changed to red. This counterfactual example is labelled as ``stop'' by the expert driver, introducing a critical decision point to the dataset. In the second scenario, the CF generator creates a more informative data point by altering the location of the front motorcycle, bringing it closer to the ego-vehicle. This modification shifts the expert's decision from ``go'' to ``stop'', allowing the imitation learner to refine its understanding of the expert's decision boundary.

As evidenced by the numerical results in Table \ref{table:per}, integrating such instances into the training dataset can help the Imitation Learner model fine-tune its policy with more samples near the expert's decision boundary. The inclusion of counterfactual examples in the training data allows the Imitation Learner model to better understand the expert driver's decision-making process in critical situations, such as those encountered near a traffic light. 

Moreover, our CF-Driver framework effectively addresses the issue of data sparsity in imitation learning, especially for rare or challenging scenarios. By generating plausible variations of the original data, we augment the training dataset with novel, yet realistic instances. This approach not only enriches the dataset but also leads to improved generalisation and overall performance of the Imitation Learner model.

The integration of counterfactual examples thus serves a dual purpose: it refines the model's decision boundaries in critical scenarios and expands the dataset to cover a broader range of driving situations. This comprehensive approach results in a more robust and accurate imitation of expert driving behaviour.

% \subsection{Quality of Counterfactual Explanations}
% \label{sec: Quality}

% Fig.~\ref{???} - Fig.~\ref{??} provide visual comparisons illustrating the perceptible disparity in the quality of 

% Finally, to elucidate how the CF examples can provide

% The above comprehensible and informative changes provided by the ?, imply that the underlying policy 

 \section{Conclusions}
 \label{section: conlusion}
 In this paper, we introduced CF-Driver, a novel end-to-end autonomous driving framework that leverages Counterfactual Explanations~(CFEs) to enhance the performance of imitation learning. By augmenting expert demonstrations with counterfactual examples, CF-Driver aims to capture a more comprehensive representation of expert behaviour and improve the robustness and generalisation capabilities of the learned driving policy.
Our evaluation results demonstrate the superior performance of CF-Driver compared to state-of-the-art CARLA drivers, showcasing the benefits of incorporating CFEs into the training process. CF-Driver achieves higher completion rates while maintaining lower collision and traffic violation rates, indicating its ability to navigate safely and efficiently in diverse driving scenarios.

Our work highlights the potential of counterfactual explanations as a valuable tool for advancing the field of autonomous driving and paves the way for further research in this direction. 
Developing more efficient methods for counterfactual generation and selection could help scale up the approach to larger datasets and more complex driving scenarios. Exploring the integration of CFEs with other learning paradigms, such as reinforcement learning or unsupervised learning, could lead to further improvements in autonomous driving performance.

\bibliographystyle{splncs04}
\bibliography{ref}

\end{document}